\documentclass[runningheads]{llncs}
\usepackage[T1]{fontenc}

\usepackage{cite}
\usepackage{graphicx}
\usepackage{wrapfig}
\usepackage{amsmath}
\usepackage{makecell}
\usepackage[dvipsnames]{xcolor}
\usepackage{todonotes}
\usepackage{ifthen}
\usepackage{float}

\usepackage[subpreambles=true]{standalone}
\usepackage{import}
\usepackage{subcaption}
\usepackage{mwe}
\usepackage{threeparttable}
\usepackage{multirow}
\usepackage{booktabs}
\usepackage{tabularx}
\usepackage{multicol}
\usepackage{multirow}
\usepackage{amsmath}
\usepackage{ragged2e}

\usepackage{placeins}  
\title{Single State Update Predictive Coding training for Time Series Forecasting and Anomaly Detection}
\titlerunning{Single State Update Predictive Coding for Time Series Forecasting}
\author{Matteo Cardoni\orcidID{0000-0002-2759-5310} \and\\ Sam Leroux\orcidID{0000-0003-3792-5026}}
\date{}
\institute{IDLab, Department of Information and Technology, Ghent University—imec,\\ 9052 Ghent, Belgium\\
\email{matteo.cardoni@ugent.be} (Corresponding author),\\ \email{sam.leroux@ugent.be}}
\begin{document}

\maketitle

%
\begin{abstract}
Predictive Coding (PC) is a neural learning paradigm that enables parallelizable neural network layer updates. However, the main bottleneck of PC Networks (PCN) is the  sequential backwards error propagation. To tackle this, we introduce a training technique that pairs a Generative PCN with a support Encoding PCN. The two PCNs are trained in parallel to match their neural activations, without sequential propagation. We apply this to time series anomaly detection and show that our approach results in more stable, continuous, online learning.
\end{abstract}
\subsubsection{Introduction and related work:} Predictive coding (PC) \cite{rao_and_ballard1999PredictiveCoding, friston2005atheory, friston2009predictive_coding_free_energy} is a learning paradigm that emerged in recent years as a low-complexity alternative for Backpropagation-based training \cite{millidge2022predictivecodingfuturedeep}. Its main attraction  is its parallelizable layer-wise updates which, in contrast to Backpropagation\cite{rumelhart1986learning}, requires no sequential layer updates.
PC training pairs each layer's activation with an additional state of the same dimension \cite{pinchetti2025benchmarkingpredictivecodingnetworks}. For each layer $l$, it involves the minimization of an energy function between the states $h_l$ and the activations $\mu_l$ \cite{pinchetti2025benchmarkingpredictivecodingnetworks,vanzwol2024predictivecodingnetworksinference}. The minimization is performed by iteratively optimizing the states and, subsequently, the weights.
Despite PC's parallelizable updates property, Predictive Coding Networks (PCN)'s layer updates require an error signal that propagates sequentially from the output throughout all the layers. Commonly, PCNs are initialized with feedforward initialization \cite{pinchetti2025benchmarkingpredictivecodingnetworks, goemaere2025erroroptimizationovercomingexponential, vanzwol2024predictivecodingnetworksinference}, which involves the states initialization with the activation values. Despite accelerating the convergence \cite{whittington2017an_approximation_of_the_error_backpropagation_algorithm_in_a_pcn, goemaere2025erroroptimizationovercomingexponential}, it requires at least as many inference steps as the network depth \cite{pinchetti2026fasterpredictivecodingnetworks, goemaere2025erroroptimizationovercomingexponential} and causes a vanishing update propagation \cite{goemaere2025erroroptimizationovercomingexponential}. To overcome the requirement for multiple state update iterations, we introduce a Guided PC training technique that pairs a Generative PCN (G-PCN) with an Encoding PCN (E-PCN), which guides and accelerates the states update by enforcing layer-wise activations matching.
Unlike prior state-update acceleration methods\cite{pinchetti2026fasterpredictivecodingnetworks, casnici2026acceleratedpredictivecodingnetworks, baskakovs2026closedformpredictivecodinghierarchical, alonso2023understandingimprovingoptimizationpredictive, salvatori2024stablefastfullyautomatic}, our solution provides parallel and hierarchical updates across the network depth, using batched data.
\vspace{-5mm}
\subsubsection{Proposed solution:}
The goal for the G-PCN, of $L$ layers, is to predict the next time-step data $x^t=h_L^t$ with $\hat x^t=\mu_L^t$. Taking inspiration from temporal Predictive Coding \cite{tang2023sequential}, the G-PCN uses $h_0^{t-1}$ to predict the same state at time $t$ ($h_0^t$), and $x^t$ from it.
Training the G-PCN and the E-PCN to produce matching activations implies the minimization of two types of energies:  the Internal Energy $\mathcal I = \frac{1}{2}\sum_l(h_l - \mu_l)^2$ between the states and the activations of the G-PCN, and the Guiding Energy $\mathcal G = \frac{1}{2}\sum_l(\mu_l - \gamma_l^2)$, between the activations of the two PCNs.

\begin{figure}[H]
    \centering
    \begin{adjustbox}{width=\textwidth}
        \includegraphics{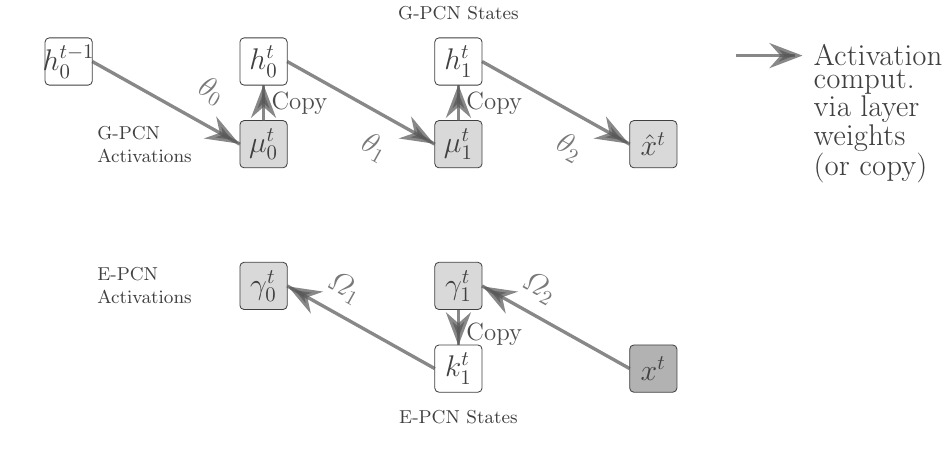}
    \end{adjustbox}
    \caption{Guided PC training step 1. Parallel feedforward initialization of the G-PCN and the E-PCN. The initializations follow  a sequential order: from $h_0^{t-1}$ to $\hat x^t$ for the G-PCN and form $x^t$ to $\gamma_0^t$ for the E-PCN. The G-PCN maps $h_0^{t-1}$ to $\mu_0^t$ via the layer $\theta_0$, that acts as temporal mapping layer.}
    \label{fig:ff_init}
\end{figure}

\begin{figure}[H]
    \centering
    \begin{adjustbox}{width=\textwidth}
        \includegraphics{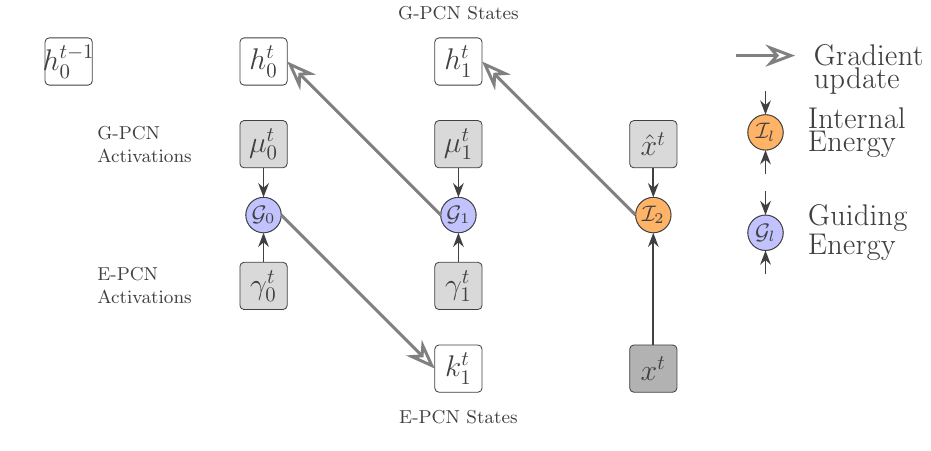}
    \end{adjustbox}
    \caption{Guided PC training step 2. G-PCN and E-PCN parallel state update, to minimize the Guiding Energy $\mathcal G$ and the Inernal Energy $\mathcal I$ at the last layer. All the the states $h_0^t, \dots, h_{L-1}^t$ and $k_{L-1}^t, \dots, k_0^t$ receive a non-zero update.}
    \label{fig:act_match_state_update}
\end{figure}

\begin{figure}[H]
    \centering
    \begin{adjustbox}{width=\textwidth}
        \includegraphics{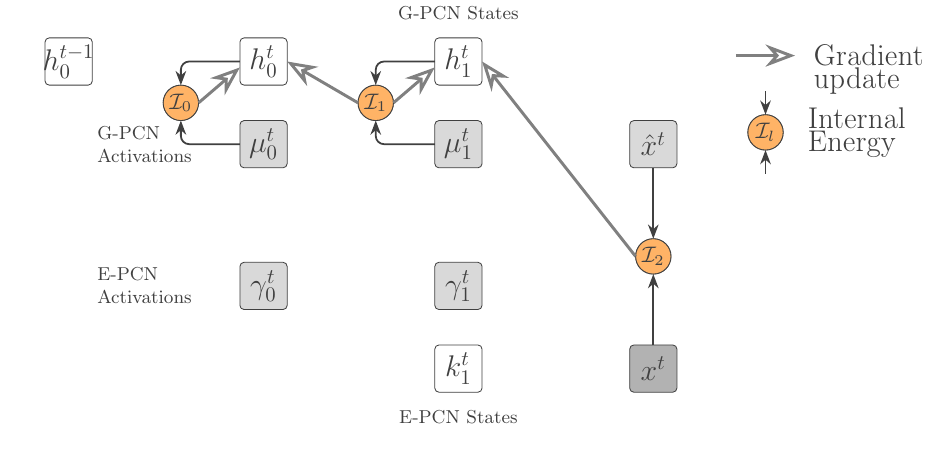}
    \end{adjustbox}
    \caption{Guided PC training step 3. G-PCN state update to minimize the Internal Energy $\mathcal I$. The states $h_0^t, \dots, h_{L-1}^t$ already received a non-zero update from step 2 (Figure \ref{fig:act_match_state_update}). Therefore, the update with respect to $\mathcal I$ will be non-zero for all $h_0^t, \dots, h_{L-1}^t$.}
    \label{fig:internal_state_update}
\end{figure}

\begin{figure}[H]
    \centering
    \begin{adjustbox}{width=\textwidth}
        \includegraphics{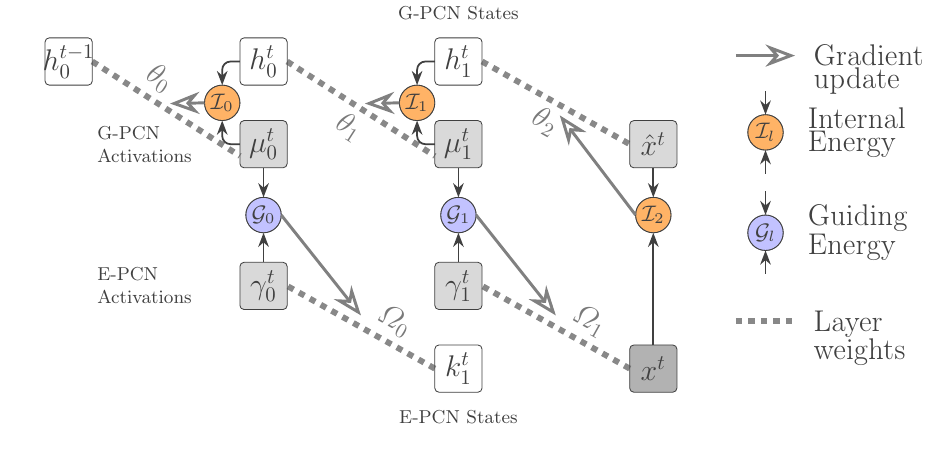}
    \end{adjustbox}
    \caption{Guided PC training step 4. G-PCN and E-PCN parallel weight update. The G-PCN weights are updated to minimize $\mathcal I$, while the E-PCN weights are updated to minimize $\mathcal G$. In this way, the E-PCN is trained to match the G-PCN representation, acting as a support network.}
    \label{fig:weight_update}
\end{figure}

For each time frame, our technique follows 4 steps. (i) First, both the G-PCN and the E-PCN perform feedforward initialization, as displayed in Figure \ref{fig:ff_init}. The G-PCN maps $h_0^{t-1}$ to $\mu_0^t$ via the layer $\theta_0$, which performs temporal mapping. The subsequent layers have the role of a PC decoder \cite{pinchetti2025benchmarkingpredictivecodingnetworks}, mapping the encoded data to $\hat x^t$.
In parallel, the E-PCN performs feedforward initialization taking $x^t$ as input, acting as an encoder.
(ii) As depicted in Figure \ref{fig:act_match_state_update}, the first part of the state update happens in the second step, where all the states of both the PCNs are updated in parallel to minimize the $\mathcal G$ and $\mathcal I_L = (\hat x^t - x^t)^2$.
(iii) The third step involves, for every layer, the classic (Vanilla) PC state update with respect to $\mathcal I $, displayed in Figure \ref{fig:internal_state_update}. It is performed only by the G-PCN as it uses $x^t$ as reference, ensuring that the E-PCN converges to the same representation. The second and third steps compose a unified,  single-iteration state update for the G-PCN.
(iv) The fourth step, that is the weight update of both the models, is displayed in Figure \ref{fig:weight_update}. The G-PCN weights are updated to minimize $\mathcal I$, as in Vanilla PC. In parallel, the E-PCN weights are updated to minimize $\mathcal G$. 

Overall, the E-PCN is trained to match the representation of the G-PCN, acting as support, similarly to a teacher model \cite{hu2022teacherstudentarchitectureknowledgelearning}. The G-PCN, instead, uses $x^t$ as reference to propagate its updates with the help of the E-PCN, similarly to a student network.
\vspace{-3mm}
\subsubsection{Experiments and results:} 
\begin{figure}
    \centering
    \begin{subfigure}{\textwidth}
       \centering
       \includegraphics{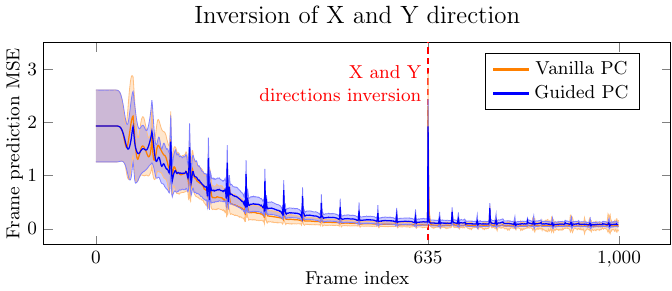}
       \caption{X and Y directions inversion at frame 635. The later MSE spikes that occur when the digit is close to that position demonstrate the adaptation to the new normality.}
    \end{subfigure}
    \begin{subfigure}{\textwidth}
        \centering
        \includegraphics{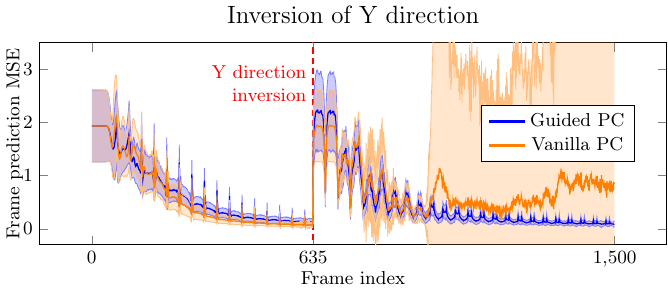}
        \caption{Y direction inversion at frame 635. The bounces become more frequent after the anomaly as the contact on the vertical and horizontal borders are not synchronous anymore.}
    \end{subfigure}
    \begin{subfigure}{\textwidth}
        \centering
        \includegraphics{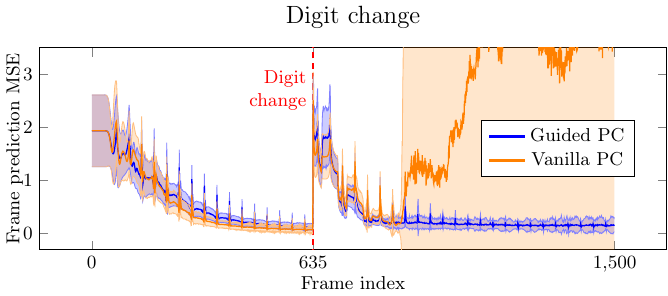}
        \caption{Digit change at frame 635, while maintaining the same direction.}
    \end{subfigure}
    \caption{Average accuracy expressed as Mean Squared Error between the ground truth images $x^t$ and the predicted images $\hat x^t$, in sequences with three different anomaly types. 100 experiments have been averaged per each category. The semi-transparent area indicates the standard deviation per time frame. The Guided PC obtains better stability with respect to anomalies, despite performing only one unified state update step}
    \label{fig:results}
\end{figure}

We apply this training technique to the task of continuous, online learning for time series anomaly detection. We created a dataset of MNIST digits \cite{mnist}  moving on a black background. We train a 5-layer MLP to predict the next frame, while batching it with previous frames. We use the Mean Squared Error (MSE) between the predicted and the actual frame as anomaly score. Starting from the initial dynamics (diagonal movement of 1 pixel), different anomalies have been introduced in the form of abrupt changes in X and/or Y direction or digit value. After an anomaly occurs, the model needs to retrain to consider the new sequence dynamics as normal. Figure \ref{fig:results} shows the anomaly scores over time. We compare traditional (Vanilla) PC training, performing 5 inference steps, with G-PCN. As the model has no prior knowledge, the anomaly scores are high at the start of training. Both techniques are capable of quickly learning the initial dynamics. The periodic spikes are caused by bouncing against the wall and changing direction. After the anomaly is introduced, the Guided PC can quickly recover to treat the new behavior as normal while the vanilla PC diverges. This is probably due to a slower and vanishing error signal propagation of Vanilla PC \cite{goemaere2025erroroptimizationovercomingexponential}.%
\vspace{-3mm}
\subsubsection{Conclusion:} We introduced a PC training technique that relies two PC models to eliminate the sequential state update bottleneck, typical of PC, and enhances training stability. Using a single unified state update step, all layers states get updated in parallel. As PC usually requires a high number of inference steps, this technique is promising in terms of edge devices online learning.


\bibliography{references}{}
\bibliographystyle{unsrt}

\section*{Appendix}
\subsubsection{Data batching:} In order to provide the PCNs with sufficient examples, every input data was composed as a batch of 128 frames $s^t = x^{t}, x^{t-1}, \dots x^{t-127}$. The training began at $t=0$ with $s^0 =x^0, \mathbf 0, \dots, \mathbf 0$. At each new time-step, every $x^t$ pushed of 1 time-step, $x^{t-127}$ is discarded and the new frame is assigned to $x^t$ . As a consequence, the G-PCN state $h_0^{t-1}$ also consists of a batch of 128 elements, each initialized to $\mathbf 0$ at $t=0$.%
\subsubsection{Training details:} To avoid excessive updates when the G-PCN already learned a good forecasting capability, an additive gaussian noise with variance $\sigma = 10^{-4}$ has been added to $h_0^{t-1}$. Moreover, the weights learning rates were modulated with respect to the ratio between the Average Mean Squared Error (MSE) between the generated frame and the input frame at time $t$ ($\text{MSE}^t$) and the same MSE at time 0 ($\text{MSE}^0$). Specifically, a Fermi-Dirac function \cite{cheol2021fermi_dirac} has been used, passing as x-value $1-\displaystyle\frac{\text{avg. MSE}^t}{\text{avg. MSE}^0}$. A learning rate $\alpha$ is modulated as $\alpha_{\text{mod}} = \frac{\alpha}{\exp\Big(\big(1-\frac{\text{avg. MSE}^t}{\text{avg. MSE}^0} - \mu\big)/KT\Big) + 1},$ using $\mu=0.9$ and $KT=0.05$.
These parameters allow the weights learning rates to stay close to their maximum value unless for a low $\text{MSE}^t$, when the weights learning rates drop to avoid undue updates.%

All experiments have been performed using the PCX framework \cite{pinchetti2025benchmarkingpredictivecodingnetworks}

\begin{table}[]
    \caption{Architectural details for the G-PCN and the E-PCN. The G-PCN forward direction goes in increasing order, while the E-PCN one goes in decreasing direction (see Figure \ref{fig:ff_init}). Every cell of the table lists the Dense layers dimensions, the layer connections and the non-linearity applied to the activation.}
    \renewcommand{\arraystretch}{1.5}
    \centering
    \begin{tabularx}{\textwidth}{
        >{\RaggedRight\arraybackslash}X
        >{\RaggedRight\arraybackslash}X
        >{\RaggedRight\arraybackslash}X
    }
    \toprule
    & \textbf{G-PCN} & \textbf{E-PCN} \\
    \midrule

    level $-1$
    & \makecell[l]{(512, 512)\\ $h_0^{t-1} \rightarrow h_0^t$\\ tanh}
    & - \\
    \midrule
    levels $0$--$2$
    & \makecell[l]{(512, 512)\\ $h_i^t \rightarrow \mu_{i+1}^t$\\ tanh}
    & \makecell[l]{(512, 512)\\ $k_{i+1}^t \rightarrow \gamma_i^t$\\ tanh} \\
    \midrule
    level $3$
    & \makecell[l]{(512, 4096)\\ $h_3^t \rightarrow \hat{x}^t$\\ linear}
    & \makecell[l]{(4096, 512)\\ $x^t \rightarrow \gamma_2^t$\\ tanh} \\

    \bottomrule
    \end{tabularx}
    \label{tab:architecure}
\end{table}
\subsubsection{Architectural details:} The architectural details for the G-PCN and the E-PCN are reported in Table \ref{tab:architecure}. Both PCNs have MLP architectures. The Vanilla PCN shares the same architecture with the G-PCN. None of the models use biases in their Dense layers, to maximize the causality between $h_0^{t-1}$ and $\hat x^t$. tanh has been used as non-linearity to keep the neural activities in a contained range, helping convergence.%
\begin{table}
    \centering
    \caption{Hyperparameters used for design space grid search. The $\beta_1$ and $\beta_2$ coefficient for the Adam optimizers have been explored in couples of the same value (using the same for both the Guided PC optimizers).}
    \setlength{\tabcolsep}{2\tabcolsep} 
    \renewcommand{\arraystretch}{2} 
    \begin{tabularx}{\textwidth}{>{\RaggedRight}p{2.5cm} >{\RaggedRight}p{3cm} >{\RaggedRight}p{3cm} >{\RaggedRight}p{3cm}}
        \toprule
                    & \makecell[l]{\textbf{SGD}\\\textbf{learning rates}}                  &  \makecell[l]{\textbf{Adam}\\\textbf{learning rates}} & \textbf{Adam} $\mathbf{\beta_1, \beta_2}$\\
        \hline
         Guided PC  & [1e-2, 5e-2, 1e-1]                                                   & [1e-4, 2.5e-4, 5e-4]                                  & \multirow{3}{*}{\makecell[l]{[(0, 0), (0.1, 0.1),\\ (0.5, 0.5), (0.9, 0.9)]}}\\[2ex]
         Vanilla PC & \makecell[l]{[1e-4, 2.5e-4, 5e-4,\\ 1e-3, 5e-3, 1e-2,\\ 5e-2, 1e-1]} & \makecell[l]{[1e-3, 5e-3, 1e-2,\\ 5e-2,  1e-1]}       & \\
        \bottomrule
    \label{tab:hyperparams_grid_search}
    \end{tabularx}
\end{table}
\subsubsection{Hyperparameters details:} Stochastic Gradient Descent was used for all the state optimizers. Adam \cite{kingma2017adammethodstochasticoptimization} was used for all the weights optimizers PCs. Table \ref{tab:hyperparams_grid_search} displays the hyperparameters used for grid search design space exploration. For the Vanilla PC, a higher number of hyperparameters was used, as only two optimizers are used (instead of 5 for the Guided PC). Table \ref{tab:hyperparams_chosen} displays the learning rates that have were to produce the results we presented. To choose the best performing configuration for Guided PC and Vanilla PC, the exploration began performing one experiment per configuration and anomaly type, using the first image of the MNIST training set. The average MSE between $x^t$ and $\hat x^t$ was evaluated. Only configurations producing the last 100 frames with $\text{average MSE}^t <= 15\% \text{ average MSE}^0$ were considered. They were subsequently ordered in order of increasing variance, as we wanted $\hat x^t$ to match $x^t$ without difference spikes. The configuration that provided the lower variance among the three anomaly types was selected and used for 100 experiments with different MNIST digits and initialization seeds, and is summarized in Table \ref{tab:hyperparams_chosen}. Both the Adam optimizers $\beta_1$ and $\beta_2$ coefficient are 0, which signifies that an absence of momentum is favorable when learning abrupt changes, as are the anomalies and bouncing.%
\begin{table}[t]
    \centering
    \caption{Learning rates used for Guided PC and Vanilla PC.}
    \renewcommand{\arraystretch}{1.5} 
    \begin{tabularx}{\textwidth}{
        *{5}{>{\RaggedRight\arraybackslash}X}
    }
        \toprule
                    & \multicolumn{2}{c}{\makecell{\textbf{Internal Energy}\\\textbf{learning rates}}} & \multicolumn{2}{c}{\makecell{\textbf{Guiding Energy}\\\textbf{learning rates}}} \\
        \cline{2-5}
                    & \textbf{States}                                & \textbf{Weights}                               & \textbf{States}                               & \textbf{Weights}                               \\
        \hline
        G-PCN       & 5e-2                                  & 2.5e-4                                & 1e-1                                 & -                                     \\
        E-PCN       & -                                     & -                                     & 1e-1                                 & 1e-4                                  \\
        Vanilla-PCN & 5e-2                                  & 2.5e-4                                & -                                    & -                                     \\
        \bottomrule
    \label{tab:hyperparams_chosen}
    \end{tabularx}
\end{table}

\begin{table}[t]
    \caption{Training time per frame averaged for 150.000 frames, from 100 experiments of 1500 frames each ($\pm$ standard deviation). The required training times are similar for the two techniques because the Guided PC implementation cannot still leverage the parallelization potential.}
    \centering
    \begin{tabular}{p{2.5cm}p{2.5cm}}
        \toprule
         \textbf{Guided PC} & \textbf{Vanilla PC}\\
         \midrule 
         $7.3 \pm 73.4$ ms & $ 7.2 \pm 85.2$ ms\\
    \bottomrule
    \end{tabular}
    \label{tab:training_times}
\end{table}
Table \ref{tab:training_times} summarizes the training times, averaged for 150.000 training frames, from 100 experiments of 1500 frames each, on NVIDIA Tesla v100-SXM3-32gb GPU. The reason why the Guided PC is does not require less time than the Vanilla PC is that, at the moment, our implementation cannot leverage on the parallelizability potential. 
\end{document}